\documentclass[conference]{IEEEtran}
\IEEEoverridecommandlockouts
\usepackage{cite}
\usepackage{amsmath,amssymb,amsfonts}
\usepackage{algorithmic}
\usepackage{graphicx}
\usepackage{textcomp}
\usepackage{xcolor}
\usepackage{algorithm}
\usepackage{subfigure}
\usepackage{makecell}
\usepackage{threeparttable}
\usepackage{multirow}
\usepackage{booktabs}
\usepackage{babel}
\IEEEoverridecommandlockouts
\usepackage[misc]{ifsym}

\newtheorem{asm}{Assumption}
\def\BibTeX{{\rm B\kern-.05em{\sc i\kern-.025em b}\kern-.08em
    T\kern-.1667em\lower.7ex\hbox{E}\kern-.125emX}}


\begin{document}

\title{LDC-VAE: A Latent Distribution Consistency Approach to Variational AutoEncoders*}

\author{\IEEEauthorblockN{Xiaoyu Chen$^{*,1}$, Chen Gong$^{*,1,2}$, Qiang He$^{1,2}$, Xinwen Hou$^{1}$ $\;$\Letter, Yu Liu$^{1}$\thanks{* THESE AUTHORS CONTRIBUTED EQUALLY.}}
\IEEEauthorblockA{$^1$ Comprehensive information system research Center, Institute of Automation, Chinese Academy of Sciences}
$^2$ School of Artificial Intelligence, University of Chinese Academy of Sciences\\
\{xiaoyu.chen, gongchen2020, heqiang2019, xinwen.hou, yu.liu\}@ia.ac.cn
}


\maketitle

\begin{abstract}
Although Variational Autoencoders (VAEs), as an important aspect of generative models, have received a lot of research interests and achieved many successful applications, it is usually a challenge to achieve the consistency between the learned latent distribution and the prior latent distribution when optimizing the evidence lower bound (ELBO), and finally leads to unsatisfactory performance in data generation. In this paper, we propose a latent distribution consistency approach to avoid such substantial inconsistency between the posterior and prior latent distributions in ELBO optimizing. We name our method as \emph{latent distribution consistency VAE (LDC-VAE)}. We achieve this purpose by assuming the real posterior distribution in latent space as a Gibbs distribution, which is approximated by an encoder network. However, there is no analytical solution for such Gibbs posterior in approximation, and traditional approximation ways are time-consuming, e.g., using the iterative sampling-based MCMC. To address this problem, we use the Stein Variational Gradient Descent (SVGD) to approximate the Gibbs posterior. Meanwhile, we use the SVGD to train a sampler net that can obtain efficient samples from the Gibbs posterior. Comparative studies on the popular image generation datasets (i.e., MNIST, CELEBA and CIFAR10) illustrate that LDC-VAE performs comparable or even better performance than several powerful improvements of VAEs. Especially, we outperform all the compared methods on MNIST with a huge margin. 
\end{abstract}


\section{Introduction}
\begin{figure*}[!t]
    \centering
    \subfigure[]{
    \includegraphics[width=0.7\linewidth]{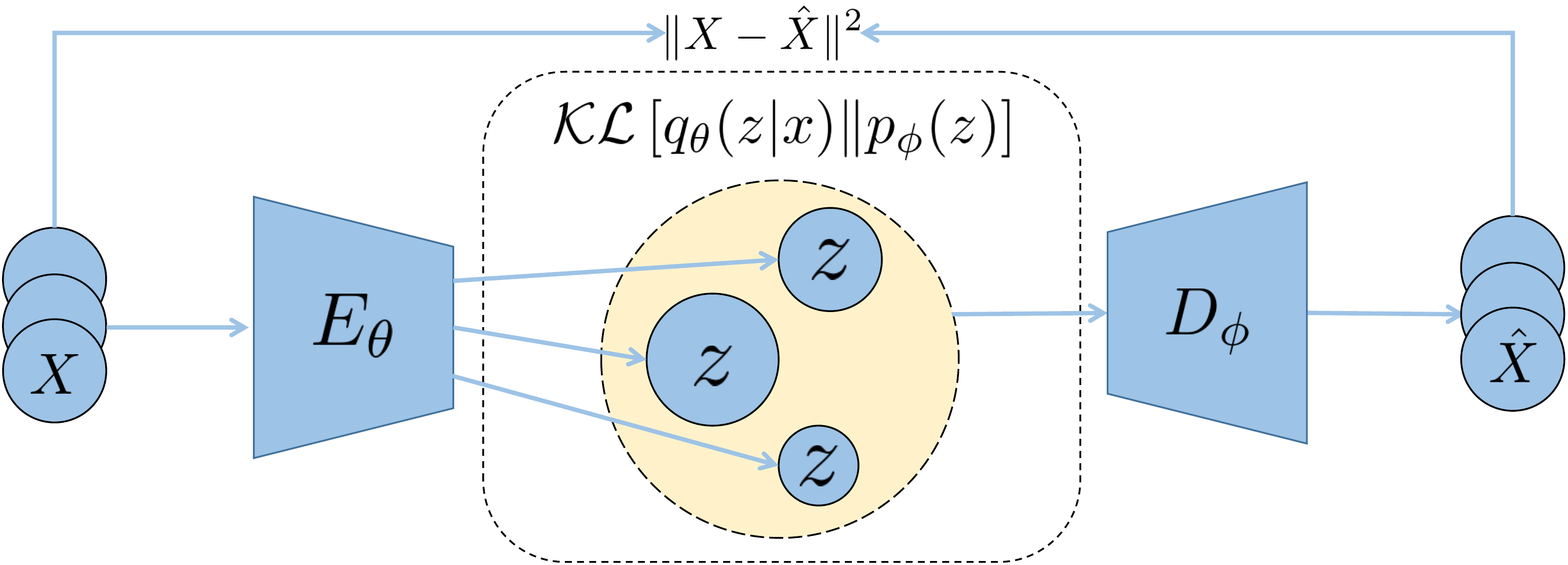}\label{graph_VAE}
    }
    \quad
    \subfigure[]{
    \includegraphics[width=0.7\linewidth]{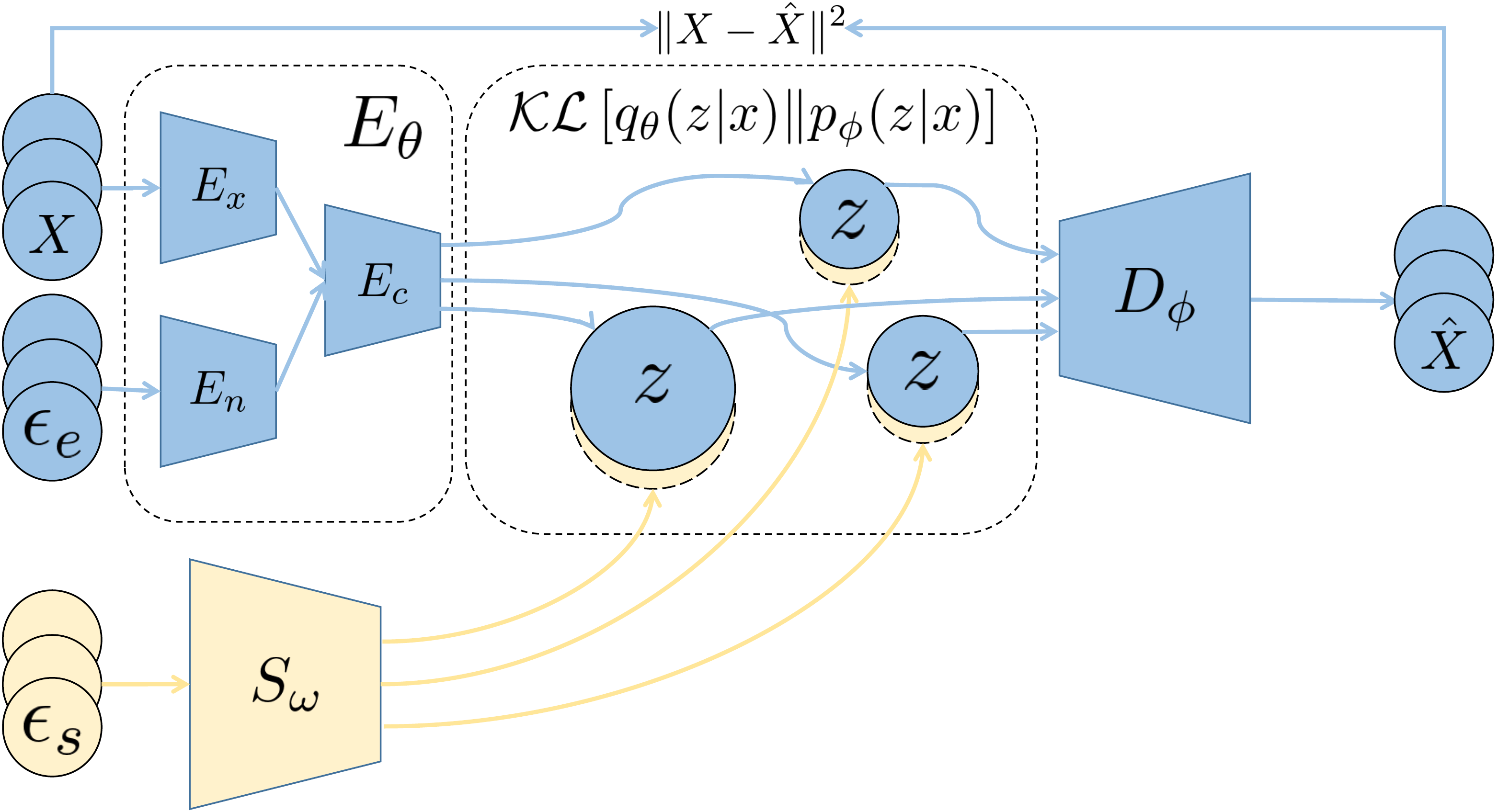}\label{graph_our}
    }
    \caption{The schematic diagram of vanilla VAE (Figure \ref{graph_VAE}) and our LDC-VAE (Figure \ref{graph_our}). In VAE, due to the substantial drawbacks in ELBO, the learned encoder distribution is always inconsistent with the prior, and leaves a lot of inconsistent regions (the uncovered yellow region in Figure \ref{graph_VAE}) that never be seen by the decoder while training. For LDC-VAE, since we use the encoder posterior to approximate the Gibbs posterior, we bypass the problems in ELBO and achieve more consistency in latent space. Since the reparameterization trick is not used in our encoder, we directly input the noise ($\epsilon_{e}$ in Figure \ref{graph_our}) from arbitrary distribution to the encoder. We add a convolutional layer ($E_{n}$) for the noise input, and leave the rest of the encoder ($E_{x}$ and $E_{c}$) identical with the RAE. Also, for efficient sampling, we train a sampler net ($S_{\omega}$) parallelly which can map the noise from arbitrary distribution to the Gibbs posterior. The sampler net has identical architecture with the encoder in RAE.}
    \label{graph_model}
\end{figure*}
In the field of machine learning, generative models, especially the unsupervised deep generative models, have been successfully applied in many fields, such as computer vision \cite{brock2018large}, natural language processing \cite{bowman2015generating}, autonomous driving \cite{sallab2017deep}, vedio games \cite{gong2021wide}, etc. With an unsupervised deep generative model, one can build the model related to the distribution on the observed data and generate more unseen data. Variational Autoencoders (VAEs) \cite{kingma2013auto} and  Generative Adversarial Networks (GANs) \cite{goodfellow2014generative}, as two of the most prominent unsupervised generative models serve the same purpose while have various differences in their theories and implementations. 

The VAEs view the data generation procedure as a random process which has a latent random variable, and utilize variational inference to train an encoder network to approximate the distribution of the latent variable. Another decoder network is trained parallelly to simulate the data generation process which can generate new data when the latent variable is given. GANs directly model the data distribution using a generator network with the input from an arbitrary distribution, and apply a discriminator to evaluate the goodness-of-fit between the generated distribution and the real data distribution. Admittedly, the architecture of GANs seems to be plainer since it models the data distribution directly, but their training process is very unstable which often leads to convergence problems \cite{arjovsky2017wasserstein,gulrajani2017improved}. In the contrast, the VAEs can provide a relatively stable training process which attributes to its solid theoretical foundation. 

However, there are several flaws of the vanilla VAE especially in the optimization of the evidence lower bound (ELBO), which weakens its performance. First of all, the Gaussian assumption of the latent space distribution gives the Kullback–Leibler (KL) divergence an analytical solution but blocks the variational inference performance \cite{burda2015importance,kingma2016improved}, since the encoder distribution is always relative with the input and hardly consistent with isotropic Gaussian. Moreover, the trade-off between consistency in the latent space and the quality of the reconstruction usually leads researchers in a dilemma. On the one hand, excessive consistency in latent space causes the sample from the posterior irrelevant to the data, making it difficult to train the decoder. On the other hand, giving the KL divergence a small coefficient indeed helps to retain more information about the input data, but it may destroy the consistency between the learned encoder distribution and the prior distribution (see Figure \ref{graph_VAE}). The samples from the inconsistent region, which between the encoder distribution (blue regions) and the prior (yellow region), may weaken the generation quality \cite{chen2016variational}.
Furthermore, the isotropic Gaussian prior is hard to achieve in the latent space substantially \cite{kingma2016improved,dai2019diagnosing,ghosh2019variational}, since the learned marginal distribution is a mixture Gaussian, which is hard to get an isotropic Gaussian marginal distribution, even all the component posteriors are relatively close to isotropic Gaussian. As the aforementioned problems of ELBO, many works focus on the improvement of the VAE\cite{burda2015importance,kingma2016improved,higgins2016beta}.

Different from the existing works which try to make improvements on the ELBO, we propose an alternative way to refine the variational inference that does not need to optimize the ELBO in VAEs. We model the gap between the real joint distribution $p_{\phi}(x,z)$ and the approximate joint distribution $q_{\theta}(x,z)$, and optimize it by approximating the posterior $p_{\phi}(z|x)$ directly rather than maximizing the ELBO (see Figure \ref{graph_our}). Our purpose approach is achieved by assuming the $p_{\phi}(z|x)$ in a Gibbs form. To realize the efficient training, we use Stein Variational Gradient Descent (SVGD) \cite{liu2016stein} to calculate the gradient of the KL divergence between the learned posterior $q_{\theta}(z|x)$ and the $p_{\phi}(z|x)$, which is inspired by Stein-VAE \cite{pu2017vae}. In the end, to address the problem which sampling from the Gibbs posterior needs the time expensively iterative method such as Markov Chain Monte Carlo (MCMC) method~\cite{andrieu2003introduction}, we train a sampler net with SVGD to make efficient sampling from the Gibbs posterior.

Our main contributions can be summarized as follows:
\begin{itemize}
    \item  We propose a novel VAE termed \emph{latent distribution consistency VAE (LDC-VAE)}. Instead of optimizing the ELBO directly, LDC-VAE uses a different way to conduct the variational inference directly under the Gibbs posterior assumption. 
    \item Sampling from the Gibbs distribution is computationally expensive or intractable for some traditional approaches (e.g., MCMC). We train a sampler net utilizing a nonparametric variational inference algorithm, i.e., SVGD, to make efficient sampling from the Gibbs posterior distribution. 
    \item The experiments illustrate that the LDC-VAE has the best FID score on MNIST among all the compared method, and has the lowest sample FID on CELEBA.
\end{itemize}

The rest parts of our paper are organized as follows. In the related works, we conclude the works which analyzed and improved the VAEs briefly. Then in the preliminaries, we retrospect the basic theories of VAEs and SVGD, which are fundamentals of LDC-VAE. Our analysis and improvement of VAEs are illustrated in the method part of this paper. In the end, we exhibit the performance of our work in both visualization form and statistics form.

\section{Related Works}
VAEs \cite{kingma2013auto} as a prominent generation model have many successful applications in various machine learning fields, such as speech synthesis \cite{luo2019learning}, 
text generation \cite{wang2019t}, transfer learning \cite{zhang2019learning}, causal effect inference\cite{louizos2017causal,yang2021causalvae,yang2020causalvae}, etc. However, the vanilla VAE usually fails in many applications. Since that, many works have been appealing in analyzing and improving VAE's theoretical framework. Some works focus on making a more complex and accurate approximation of the posterior over the latent space to better fit the prior, such as using several samples to learn richer posterior \cite{burda2015importance}, or adding an Inverse
Autoregressive Flow (IAF) after the encoder\cite{kingma2016improved}. 

Although these works can make more flexible and powerful approximation of the posterior in the latent space than the vanilla VAE, the optimization objective of the encoder and decoder still remain the primitive problems. In fact, the KL divergence that forces every posterior close to the prior distribution is equivalent to making the posterior irrelevant to the input data \cite{chen2016variational,zhao2017towards}. However, the loss of the decoder needs the information related to the data to ensure the quality of the reconstruction \cite{tolstikhin2017wasserstein}. To solve this problem and get relative balance between the distribution consistency and reconstruction quality, researchers propose a series of methods\cite{bowman2015generating,higgins2016beta}.

But even we apply all these methods to find an optimal balance between the KL divergence and reconstruction loss, the gap between the distribution of encoding space and the isotropic Gaussian prior still exists \cite{kingma2016improved,dai2019diagnosing}. This gap leads to the problem that some samples in latent space from the region never be seen by the decoder in training, and finally causes a quality loss in generating new data. Up to present, there are several ways to fix this notorious issue, such as choosing an autoregressive prior which is obtained by the IAF with the random noise \cite{chen2016variational}; apply a prior computing by a mixture posterior distribution conditioned on the learnable pseudo-inputs \cite{tomczak2018vae}. Besides, it is a natural intuition that utilizing the Gaussian mixture model is a straightforward method, but it is not flexible enough comparing to the method above.

Our work is different with these aforementioned methods that we do not make improvement on the ELBO. We bypass it using the Gibbs posterior assumption.

\section{Preliminaries}
\subsection{Variational AutoEncoder}
VAEs model the data generation process as a random process with latent variable $z$, and try to inference its posterior distribution $p_{\phi}(z|x)$. While this posterior seems to be intractable, VAEs use an approximate distribution $q_{\theta}(z|x)$ to approach it:
\begin{equation}\label{eq:VAE_1}
    \text{KL} \left[ q_{\theta}(z|x) \| p_{\phi}(z|x) \right] = - \text{ELBO}\left( \theta,\phi,x \right) + \log p_{\phi}(x)
\end{equation}

Since the $\log p_{\phi}(x)$ is fixed when the dataset is given, minimizing the KL divergence on the LHS of Eq. (\ref{eq:VAE_1}) is equivalent to maximizing the ELBO:
\begin{equation}\label{eq:ELBO}
    \begin{aligned}
     &\text{ELBO}\left( \theta,\phi,x \right) \\ 
     & \qquad = -\text{KL} \left[ q_{\theta}(z|x) \| p_{\phi}(z) \right] + \mathbb{E}_{q_{\theta}(z|x)} \left[ \log p_{\phi}(x|z) \right] 
    \end{aligned}
\end{equation}

To implement VAEs for data generation, the parametric forms of the three distribution in Eq. (\ref{eq:ELBO}) must be specified. For simplicity, most of VAEs use the Gaussian form, where the $p_{\phi}(z) \sim \mathcal{N}(0,I) $. The $q_{\theta}(z|x)$and $p_{\phi}(x|z)$ are assumed to be the Gaussian distribution which parameter is determined by the output of the encoder and the decoder respectively:
\begin{equation}\label{eq:VAE_3}
    \begin{gathered}
        z \sim q_{\theta}(z|x) = \mathcal{N}(z;\mu_{\theta}(x), \sigma_{\theta}^{2}(x)I)
    \end{gathered}
\end{equation}
\begin{equation}\label{eq:VAE_4}
    \begin{gathered}
        x \sim p_{\phi}(x|z) = \mathcal{N}(x;\mu_{\phi}(z), \sigma_{\phi}^{2}(z)I)
    \end{gathered}
\end{equation}

Based on the Guassian assumption, the KL divergence term in Eq. (\ref{eq:ELBO}) is integrated analytically and calculated with the reparameterization trick. When the dimension of $z$ is $d$, the $\mathcal{L}_{KL}$ loss is formulated as follow:
\begin{equation}
    \begin{gathered}
        \mathcal{L}_{KL} = \frac{1}{2} \sum\nolimits_{j=1}^{d} \left[ \mu_{\theta}^{j}(x)^2 + \sigma_{\theta}^{j}(x)^2 - \log\left(\sigma_{\theta}^{j}(x)^2\right) - 1 \right]
    \end{gathered}
\end{equation}

For the reconstruction, the value of the expectation of log-likelihood is approximated by Monte Carlo method which costs lots of computing resource for accuracy:
\begin{equation}
    \begin{gathered}
         \mathbb{E}_{q_{\theta}(z|x)} \left[ \log p_{\phi}(x|z) \right] \approx \frac{1}{M} \sum\limits_{m=1}^{M} \log p_{\phi}(x|z_{m})
    \end{gathered}
\end{equation}

However, experiments in vanilla VAE  show that the $M$ can be set to 1 when the batch size of $x$ is big enough (e.g. 100 in their paper)~\cite{kingma2013auto}. Therefore, for simplicity, set the $\sigma_{\phi}^{2}(z)I = I$ and let $M=1$, we have the reconstruction loss,
\begin{equation}
    \begin{gathered}
        \mathcal{L}_{REC} = \|x-\mu_{\phi}(z)\|^2
    \end{gathered}
\end{equation}

\subsection{Stein Variational Gradient Descent}
In this section, we introduce the basic of SVGD, which is a nonparametric variational inference algorithm that integrates ideas from Stein method, kernel method and variarional inference \cite{liu2016kernelized,liu2016stein,liu2018stein}. Recent years have witnessed the success of SVGD in machine learning \cite{haarnoja2018soft,wang2018stein,gong2020stable}.
\begin{asm}
We assume that $q(x): \mathcal{X} \subset \mathbb{R}^n$ is a continuously differentiable and positive p.d.f, and $\boldsymbol{\Phi}(x): \mathbb{R}^n \to \mathbb{R}^n$ equals to $\left[\phi_{1}(x), \cdots, \phi_{d}(x)\right]^{\top}$, which is a smooth vector function.
\end{asm}
The SVGD aims to transport a set of inital particles $\{x_i\}_{i=1}^n$ to approximate the given target posterior distributions $p(x)$. The particles set $\{x_i\}_{i=1}^n$ is sampled from the distribution $q(x)$. SVGD achieves the approximation by leveraging efficient deterministic dynamics to iteratively updating the inital particles set $\{x_i\}_{i=1}^n: x_i \gets x_i + \epsilon \boldsymbol{\Phi}^\ast(x_i)$, where $\epsilon$ is a small step size and $\boldsymbol{\Phi}^\ast$ is a function chosen to maximally decrease the KL divergence. The $\boldsymbol{\Phi}^\ast$ is
\begin{equation}
     \boldsymbol{\Phi}^{*}=\underset{\boldsymbol{\Phi} \in \mathcal{B}}{\arg \max } \lim_{\epsilon \to 0}\left\{-\frac{\mathrm{d}}{\mathrm{d} \epsilon} \mathrm{KL}\left(q_{[\epsilon \boldsymbol{\Phi}]}(x) \| p(x)\right)\right\},
\label{eq:stein}
\end{equation}

\noindent where $q_{[\epsilon \boldsymbol{\Phi}]}$ is the particles distribution after the update leveraging $x \gets x + \epsilon \boldsymbol{\Phi}(x)$, and $x \sim q(x)$; $\mathcal{B}$ denotes the unit ball of reproducing kernel Hilbert space (RKHS): $\mathcal{H}^d := \mathcal{H}_0 \times \mathcal{H}_0 \cdots \mathcal{H}_0$, and $\mathcal{B} = \{\boldsymbol{\Phi} \in \mathcal{H}^{d} \mid \|\boldsymbol{\Phi}\|_{\mathcal{H}^{d}} \leq 1\}$. Fortunately, Liu \& Wang prove that the result of Eq. (\ref{eq:stein}) can be transformed to a linear function of $\boldsymbol{\Phi}$ \cite{liu2016stein},
\begin{equation}
\begin{gathered}
    \lim_{\epsilon \to 0}-\frac{\mathrm{d}}{\mathrm{d} \epsilon} \operatorname{KL}\left(q_{[\epsilon \boldsymbol{\Phi}]}(x) \| p(x)\right)=\mathbb{E}_{x \sim q(x)}\left[\operatorname{trace}\left(\mathcal{T}_{p}^{\top} \boldsymbol{\Phi(x)}\right)\right] \\ 
    \mathcal{T}_{p}^{\top} \boldsymbol{\Phi}(x)=\nabla_{x} \log p(x)^{\top} \boldsymbol{\Phi}(x)+\nabla_{x}^{\top} \boldsymbol{\Phi}(x),
\end{gathered}
\label{eq:stein_1}
\end{equation}

\noindent where $\mathcal{T}_p$ is called \emph{Stein operator}. The $\mathcal{T}_p$ and derivative $\nabla_x$ is considered as $\mathbb{R}^n$ column vectors, so the $\mathcal{T}_{p}^{\top} \boldsymbol{\Phi}(x)$ and $\nabla_{x}^{\top} \boldsymbol{\Phi}(x)$ can be viewed as inner products, e.g., $\nabla_{x}^{\top} \boldsymbol{\Phi}(x) = \sum_{j=1}^d \nabla_{x^j} \phi_j(x) = \langle\nabla_{x}^{\top}, \boldsymbol{\Phi} \rangle$, where $x^j$ and $\phi_j$ are the $j$-th variable of vector $x$ and $\boldsymbol{\Phi}$. As the $\boldsymbol{\Phi} \in \mathcal{H}^{d}$, the Eq. (\ref{eq:stein_1}) equals to
\begin{equation}
    \begin{aligned}
        \boldsymbol{\Phi}_{q,p}^\ast(\cdot) = \mathbb{E}_{x\sim q(x)} \left[ \mathcal{K}(x,\cdot) \nabla_x \log p(x) + \nabla_x \mathcal{K}(x,\cdot)\right],
    \end{aligned}
\end{equation}

\noindent where $\mathcal{K}(x,\cdot)$ is a positive define kernel associated with RKHS $\mathcal{H}^d$. We obtain the Stein variational gradient to approximate $\{x_i\}_{i=1}^n \sim q(x)$ to $p(x)$.

\section{Method}
Although the theoretical framework gives VAEs the merit of implementation convenience and training stability, the intrinsic drawbacks hinder VAEs to perform a satisfying generation result. Actually, as we mentioned above, the ELBO plays an important role in these defects.
In this section, we introduce our method, LDC-VAE, which can be seen as an alternative version of VAEs without the ELBO.

\subsection{The LDC-VAE}
To solve the inconsistency in $\mathcal{L}_{KL}$ and the conflict between $\mathcal{L}_{KL}$ and  $\mathcal{L}_{REC}$ in Eq. (\ref{eq:ELBO}), we use a different formula that gets rid of the problems in the ELBO. We achieve this by minimizing the KL divergence between the joint distribution $q_\theta(x,z)$ and $p_\phi(x,z)$, which is equivalent to minimize the KL divergence of two posterior distribution:
\begin{equation}\label{eq:method_1}
    \begin{aligned}
        &\text{KL}\left[q_{\theta}(x, z) \| p_{\phi}(x, z)\right] \\
        &\qquad\qquad= \iint q_{\theta}(x, z) \log \frac{q_{\theta}(x, z)}{p_{\phi}(x, z)} d z d x \\
        &\qquad\qquad= \iint q_{\theta}(x)q_{\theta}(z|x) \log \frac{q_{\theta}(x)q_{\theta}(z|x)}{p_{\phi}(x)p_{\phi}(z|x)} d z d x \\
        &\qquad\qquad= \iint q_{\theta}(x)q_{\theta}(z|x) \log \frac{q_{\theta}(z|x)}{p_{\phi}(z|x)} d z d x \\
        &\qquad\qquad= \mathbb{E}_{x\sim q_{\theta}(x)}\left[ \text{KL}\left[ q_{\theta}(z|x) \| p_{\phi}(z|x) \right]\right]
    \end{aligned}
\end{equation}

In the Eq. (\ref{eq:method_1}), we assume that $p_{\phi}(x)=q_{\theta}(x)$, since the $x$ in the $q_{\theta}(z|x)$ and the $p_{\phi}(z|x)$ both represent the data from the same dataset. The $p_{\phi}(z|x)$ is the true posterior distribution of the latent variable $z$, and $q_{\theta}(z|x)$ is our approximation posterior distribution which is approximated by a deep neural network. Besides, the KL loss in our model is as follows:
\begin{equation}
    \begin{gathered}
        \mathcal{L}_{\theta} = \text{KL}\left[ q_{\theta}(z|x) \| p_{\phi}(z|x) \right]
    \end{gathered}
\end{equation}

For the true posterior $p_{\phi}(z|x)$, we assume it has a Gibbs form,
\begin{equation}\label{Gibbs}
    \begin{gathered}
        p_{\phi}(z|x) = \frac{1}{C} \exp{\left\{-\frac{\|x-D_{\phi}(z)\|^{2}}{\sigma^2}\right\}}
    \end{gathered}
\end{equation}

The $D_{\phi}(z)$ represents the reconstruction data. Although this formula does not seem to be a posterior distribution of $z$, we must point out that for every given $x$, the Eq. (\ref{Gibbs}) is a function of $z$, and can satisfy the equation $\int p_{\phi}(z|x) dz = 1$ with an appropriate normalization factor C. Different with vallina VAEs, we do not assume the form of $q_{\theta}(z|x)$, since the assumption weakens the flexibility of learned posterior distribution.

In the training, we use the encoder to approximate the Gibbs posterior (i.e., in Eq. (\ref{Gibbs})), and the learned latent distribution $q_{\theta}(z) = \int q_{\theta}(z|x)q_{\theta}(x) dz = \int q_{\theta}(z|x)p_{\phi}(x) dz $ is finally consistent with the $p_{\phi}(z) = \int p_{\phi}(z|x) p_{\phi}(x) dx$ if the training process convergence properly. Since we use a posterior to approximate the another posterior, the consistency in the formula acts as an insurance for the goodness-of-fit, helping our algorithm gets rid of the defects in the ELBO. 

For the reconstruction loss, we minimize the $\mathcal{L}_{REC} = \|x-D_{\phi}(z)\|^2$. It is equivalent to maximize the log-likelihood of the $z$ from the Gibbs posterior when $x$ has been observed, which is the second RHS term in the last equation of Eq. (\ref{eq:recon}). 

\begin{equation}\label{eq:recon}
    \begin{aligned}
        &\mathbb{E}_{x\sim q_{\theta}(x)}\left[ \text{KL}\left[ q_{\theta}(z|x) \| p_{\phi}(z|x) \right]\right] \\
        &\qquad= \mathbb{E}_{x\sim q_{\theta}(x)} \left[ \int q_{\theta}(z|x) \log\frac{q_{\theta}(z|x)}{p_{\phi}(z|x)} dz \right] \\
        &\qquad= \mathbb{E}_{x\sim q_{\theta}(x), z\sim q_{\theta}(z|x)} \left[ \log q_{\theta}(z|x) \right] \\
        &\qquad- \mathbb{E}_{x\sim q_{\theta}(x), z\sim q_{\theta}(z|x)} \left[ \log p_{\phi}(z|x) \right]
    \end{aligned}
\end{equation}

The $\mathcal{L}_{\theta}$ and $\mathcal{L}_{REC}$ in our model are interpreted as follows: minimizing the $\mathcal{L}_{REC}$ loss will maximizing the likelihood of $z \sim p_{\phi}(z|x)$,  and minimizing the $\mathcal{L}_{\theta}$ can make our encoder distribution consistent with $p_{\phi}(z|x)$.

Regarding the Gibbs posterior including the information about $x$, the $\mathcal{L}_{\theta}$ and $\mathcal{L}_{REC}$ in LDC-VAE are not mutually exclusive as what happened in the ELBO. Both the minimization of these two loss will lead to the minimization of the KL between the two joint distributions, so they can be optimized without conflict. 

For the flexibility of the encoder distribution, we use a different implementation of encoder which does not apply the reparameterization trick. We input the noise sampled from an arbitrary distribution to the encoder and obtain the sample result directly (see Figure \ref{graph_our}). The output of our encoder is random even the input $x$ is fixed: $z = E_{\theta}(x, \epsilon_{e}) \sim q_{\theta}(z|x)$. While in VAEs, the output of encoder represent the $\mu_{\theta}(x)$ and $\sigma^{2}_{\theta}(x)$ of the Gaussian distribution and will be fixed when the $x$ is given. Due to this adaption, our encoder posterior distribution can approximate the arbitrary distribution which depends on the parameter of the encoder.

\begin{algorithm}[!t]
\caption{The LDC-VAE training process in iteration $t$}
\label{alg:training process}
\textbf{Input}:\\
Batch of data $x \sim p_{\phi}(x)$;\\
Batch of noise from arbitrary distribution for the encoder $\epsilon_{e} \sim f_{e}(\epsilon)$;\\
Batch of noise from arbitrary distribution for the sampler net $\epsilon_{s} \sim f_{s}(\epsilon)$;\\
Model parameters of encoder and decoder $E_{\theta}^{(t)}$, $D_{\phi}^{(t)}$;\\
Model parameters of sampler net $S_{\omega}^{(t)}$\\
\textbf{Output}:\leftline{Updated parameters: $E_{\theta}^{(t+1)}$,$D_{\phi}^{(t+1)}$, $S_{\omega}^{(t+1)}$}
\begin{algorithmic}[1] 
\STATE Obtain the output of encoder and decoder:\\ $z_{\theta} = E_{\theta}^{(t)}(x,\epsilon_{e})$, $\hat{x} = D_{\phi}^{(t)}(z_{\theta})$.
\STATE Compute the reconstruction loss:\\ $\mathcal{L}_{REC} = MSE(x,\hat{x})$
\STATE Calculate the gradient of encoder:\\
$\Phi^{\ast}_{\theta}(z_{\theta}) = \frac{1}{n}\sum\limits_{i=1}^{n} \mathcal{K}(z_{i},z_{\theta}) \cdot \nabla_{z_{i}}\frac{\mathcal{L}_{REC}}{\sigma^2} + \nabla_{z_{i}} \mathcal{K}(z_{i},z_{\theta})$,\\
$\frac{\partial \mathcal{L}_{\theta}}{\partial \theta} =  \Phi^{\ast}_{\theta}(z_{\theta}) \cdot \frac{\partial z_{\theta}}{\partial \theta}$
\STATE Update the parameter of encoder and decoder:\\
$E_{\theta}^{(t+1)}$, $D_{\phi}^{(t+1)}$ = Adam($E_{\theta}^{(t)}$, $D_{\phi}^{(t)}$, $\frac{\partial \mathcal{L}_{\theta}}{\partial \theta}$, $\frac{\partial \mathcal{L}_{REC}}{\partial \phi}$)
\STATE Obtain the output of sampler net:\\ $z_{\omega} = S_{\omega}^{(t)}(\epsilon_{s})$
\STATE Calculate the gradient of sampler net:\\
$\Phi^{\ast}_{\omega}(z_{\omega})=\frac{1}{n}\sum\limits_{i=1}^{n} \mathcal{K}(z_{i},z_{\omega}) \cdot \nabla_{z_{i}}\frac{\mathcal{L}_{REC}}{\sigma^2} + \nabla_{z_{i}} \mathcal{K}(z_{i},z_{\omega})$,\\
$\frac{\partial \mathcal{L}_{\omega}}{\partial \omega} = \Phi^{\ast}_{\omega}(z_{\omega}) \cdot \frac{\partial z_{\omega}}{\partial \omega}$
\STATE Update the parameter of sampler net:\\
$S_{\omega}^{(t+1)}$ = Adam($S_{\omega}^{(t)}$, $\frac{\partial \mathcal{L}_{\omega}}{\partial \omega}$)
\end{algorithmic}
\end{algorithm}
\subsection{Optimization with SVGD}
Although the $\mathcal{L}_{\theta} = \text{KL}\left[ q_{\theta}(z|x) \| p_{\phi}(z|x)  \right]$ can be approximated using sampling-based method such as MCMC, it is too slow and has poor convergence. In our implementation, we use the SVGD to approximate the gradient of $\mathcal{L}_{\theta}$ w.r.t. the parameter $\theta$.

Specifically, we first compute the gradient of $\mathcal{L}_{\theta}$ w.r.t. the $z_{\theta}$:
\begin{equation}
    \begin{aligned}
        &\Phi^{\ast}_{\theta}(z_{\theta}) \\
        &\quad= \mathbb{E}_{z_i \sim q_{\theta}(z|x)} \left[ \mathcal{K}(z_{i},z_{\theta})\nabla_{z_{i}}\log p_{\phi}(z|x) + \nabla_{z_{i}} \mathcal{K}(z_{i},z_{\theta}) \right] \\
        &\quad= \frac{1}{n}\sum\limits_{i = 1}^{n} \left[\mathcal{K}(z_{i},z_{\theta}) \cdot \nabla_{z_{i}}\frac{\mathcal{L}_{REC}}{\sigma^2} + \nabla_{z_{i}} \mathcal{K}(z_{i},z_{\theta}) \right]
    \end{aligned}
\end{equation}

Then we set $\Phi^{\ast}_{\theta}(z_{\theta})$ act as a constant in the backpropagation and obtain the gradient of KL divergence w.r.t. to the parameter of encoder according to the chain rule:

\begin{equation}
    \begin{aligned}
        \frac{\partial \text{KL}\left[ q_{\theta}(z|x) \| p_{\phi}(z|x) \right]}{\partial \theta} = &  \frac{\partial \text{KL}\left[ q_{\theta}(z|x) \| p_{\phi}(z|x) \right]}{\partial z_{\theta}} \frac{\partial z_{\theta}}{\partial \theta} \\
    = & \Phi^{\ast}_{\theta}(z_{\theta}) \frac{\partial z_{\theta}}{\partial \theta}\\
    \end{aligned}
\end{equation}

For the purpose that efficiently samples from the $p_{\phi}(z|x)$, we train a sampler net $S_{\omega}(\epsilon_{s})$ which maps an arbitrary distribution $\epsilon_{s} \sim f_{s}(\epsilon)$ to the $p_{\phi}(z|x)$ using SVGD. The gradient of KL divergence w.r.t to the parameter of sampler net has the same formula as the encoder. We present Algorithm \ref{alg:training process}, depicting the process of utilizing the SVGD to train a VAE.

\section{Experiment}
\begin{figure*}[t]
    \centering
    \subfigure[]{
    \includegraphics[width=0.45\linewidth]{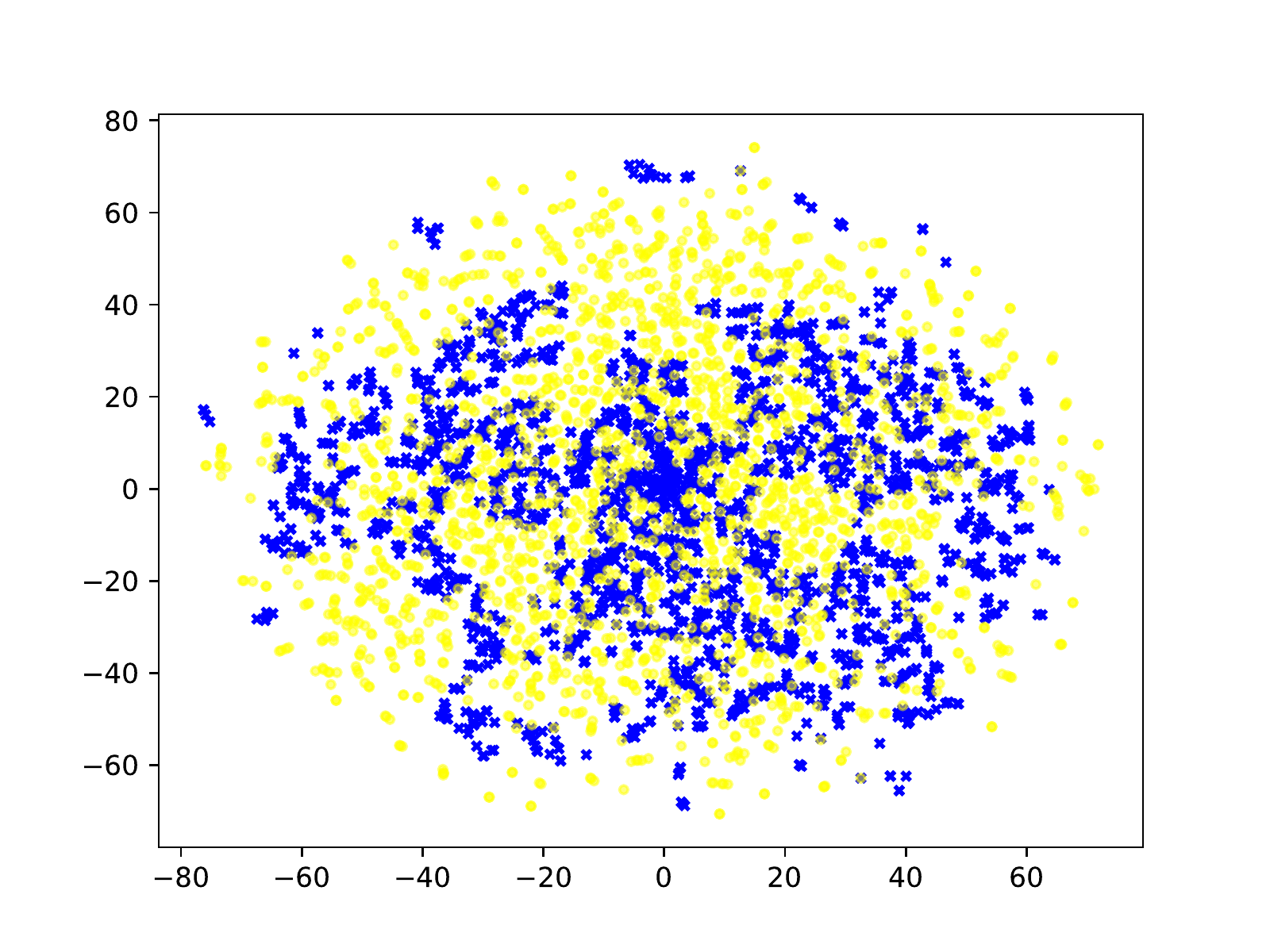}\label{latent_VAE}
    }
    \quad
    \subfigure[]{
    \includegraphics[width=0.45\linewidth]{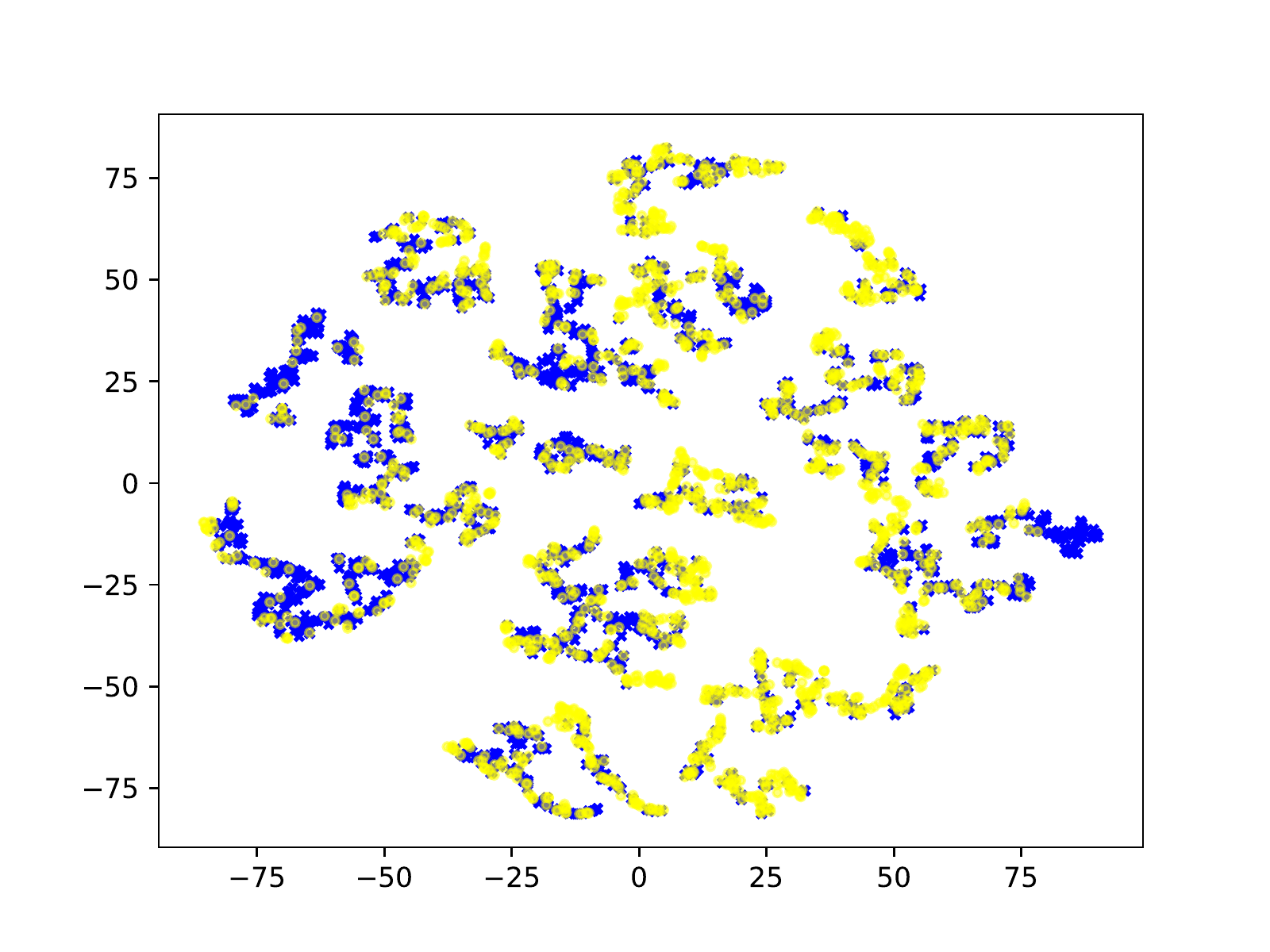}\label{latent_our}
    }
    \caption{Visualizing the latent space. The 16-dimensional samples are project to 2-dimensional using T-SNE. The blue points are encoder samples, while the yellow points come from the Gaussian prior in Figure \ref{latent_VAE} and Gibbs posterior in Figure \ref{latent_our}. It is obvious that the inconsistency between the encoder distribution and isotropic Gaussian. And clearly, our method fix this problem to get a consistent latent space between the training and generation.}
    \label{vis_latent}
\end{figure*}

\begin{table*}[t]
\caption{Quantitative evaluation of models using FID score.}
\centering
\begin{center}
\begin{tabular}{p{60 pt}<{\centering}|p{35 pt}<{\centering}|p{35 pt}<{\centering}|p{35 pt}<{\centering}|p{35 pt}<{\centering}|p{35 pt}<{\centering}|p{35 pt}<{\centering}|p{35 pt}<{\centering}|p{35 pt}<{\centering}|p{35 pt}<{\centering}}
\hline
\multirow{2}{*}{Models} &\multicolumn{3}{c}{\textbf{MNIST}}&\multicolumn{3}{|c|}{\textbf{CELEBA}}&\multicolumn{3}{|c}{\textbf{CIFAR10}} \\
\cline{2-10}
 & \textbf{\textit{Rec$^{\mathrm{a}}$}}& \textbf{\textit{Sample$^{\mathrm{b}}$}} & \textbf{\textit{Interp$^{\mathrm{c}}$}} & \textbf{\textit{Rec}} & \textbf{\textit{Sample}} &\textbf{\textit{Interp}} & \textbf{\textit{Rec}} & \textbf{\textit{Sample}} & \textbf{\textit{Interp}}\\
\hline
    VAE & 18.26 & 19.21 & 18.21 & 39.12 & 48.12 & 44.49 & 57.94 & 106.37 & 88.62 \\
    CV-VAE & 15.15 & 33.79 & 25.12 & 40.41 & 48.87 & 44.96 & 37.74 & 94.75 & 69.71\\
    WAE-MMD$^{\mathrm{d}}$ & 10.03 & 20.42 & 14.34 & \bf{34.81} & 53.67 & 40.93 & 35.97 & 117.44 & 76.89 \\
    2sAE & 20.31 & 18.81 & 18.35 & 42.04 & 49.70 & 47.54 & 62.54 & 109.77 & 89.06 \\
    RAE-L2$^{\mathrm{e}}$ & 10.53 & 8.69 & 14.54 & 43.52 & 47.97 & 45.98 & 32.24 & \bf{74.16} & \bf{62.54} \\
    RAE-SN$^{\mathrm{g}}$ & 15.65 & 11.74 & 15.15 & 36.01 & 40.95 & \bf{39.53} & \bf{27.61} & 75.30 & 63.62 \\
    RAE-GP$^{\mathrm{f}}$ & 14.04 & 11.54 & 15.32 & 39.71 & 45.63 & 47.00 & 32.17 & 76.33 & 64.08 \\
     \hline
    LDC-VAE & \bf{1.02} & \bf{4.73} & \bf{8.09} & 34.82 & \bf{36.84} & 44.27 & 49.18 & 101.24 & 88.41 \\
\cline{1-10}
\end{tabular}
\begin{tablenotes}
       \footnotesize
        \item[1] All the models are evaluated by $^{\mathrm{a}}$Rec:reconstruct the image from test set; $^{\mathrm{b}}$Sample:generating new image using samples in latent space; $^{\mathrm{c}}$Interp:generating new image using the mean of two random pair in the latent space from the test set. For the compared model, $^{\mathrm{d}}$WAE-MMD denote the WAE with MMD loss. $^{\mathrm{e}}$RAE-L2 denote the RAE with L2 regularization, $^{\mathrm{f}}$GP denote the gradient penalty and $^{\mathrm{g}}$SN denote the spectral normalization.
        All the sample FID of RAE models refers to generate new images using the ex-post density estimation (Gaussian Mixture Model).
     \end{tablenotes}
\end{center}
\label{FID score}
\end{table*}

To evaluate our method, we design our experiment to test our model in two aspects: (i) the consistency of distribution in the latent space, (ii) reconstruction, sample and interpolation quality. In this section, we firstly describe our experiment details. Then, we provide the visualized evidence to show the inconsistency of VAEs and how LDC-VAE achieves consistency in latent space. At the end of this section, we evaluate the image generation performance of our LDC-VAE in quantitive and qualitative ways.

\subsection{Experimental Setup}
In all the experiments on three datasets, we set $f_{e}(\epsilon) = f_{s}(\epsilon) = \mathcal{N}(0,I)$ for the encoder and the sampler net. For the Gibbs posterior, we set the $\sigma^{2}$ as the variance of $\|X-D_{\phi}(z)\|$ in a batch. Due to the application of the SVGD, the normalization factor C is trivial in our method. For the kernel function, we use the RBF kernel function $\mathcal{K}(z, z_{i}) = \exp \left(-\frac{1}{h} ||z-z_{i}||^2\right) $. We set bandwith $h = \text{med}^2/\log n$, with med equal to the median of all the pairwise distance between $z_{i}$ for all the $n$ samples. According to \cite{liu2016stein}, this setting can balance the deviation from other $z_{j}$ while computing the gradient of $z_{i}$. For the network architecture, we adapt the same network as that in RAE \cite{ghosh2019variational}. We add a convolutional laryer for the noise input in the encoder, and set our sampler net as same as the encoder in the RAE. For the convolution kernel size, we use the same kernel size in RAE, which has $4\times 4$ kernel size for MNIST and CIFAR-10, and $5\times 5$ for CELEBA. Also, we use a batchsize $=100$ and $[16,64,128]-$dimensional latent space for MNIST, CELEBA and CIFAR10, which are identical to the 
hyperparameters in RAE. All the parameters of our networks are optimized using Adam \cite{kingma2014adam}.

\subsection{Visualizing the Consistency of Distribution}
The inconsistency between the encoder distribution and the prior is a vital defect of VAEs. This problem causes some samples from the prior came from the region that never be seen by the decoder, and finally leads to the poor generation quality.
To give clear visual evidence of this problem and prove that our method indeed alleviate this phenomenon, we train both the vanilla VAE and our model on MNIST with 16-dimensional latent space. To visualize the inconsistency problem of VAEs, we compare 2000 test set samples from the encoder and 2000 samples from the prior. Besides, to prove the consistency in latent space of our model, we compare the test set samples with samples from our sampler net. Both of the comparison are using scatter plot and project the 16-dimensional latent space to 2-dimensional using T-SNE \cite{van2008visualizing}.

As it can be seen in Figure \ref{latent_VAE}, the marginal distribution $q_{\theta}(z) = \int q_{\theta}(z|x) p_{\phi}(x) dx$ is not consistent well to the isotropic Gaussian and left a lot of uncovered regions. When generating new data, samples from these regions may have poor generation quality, since the decoder has never been trained to decode the latent variable from these regions. But with our model, due to the application of Gibbs distribution and our sampler net, the distributions in latent space can be much more consistent.

\subsection{Quantitive and Qualitative Result for Image Generation}
\begin{figure*}[!t]
    \centering
    \subfigure[MNIST reconstruction.]{
    \includegraphics[width=0.31\linewidth]{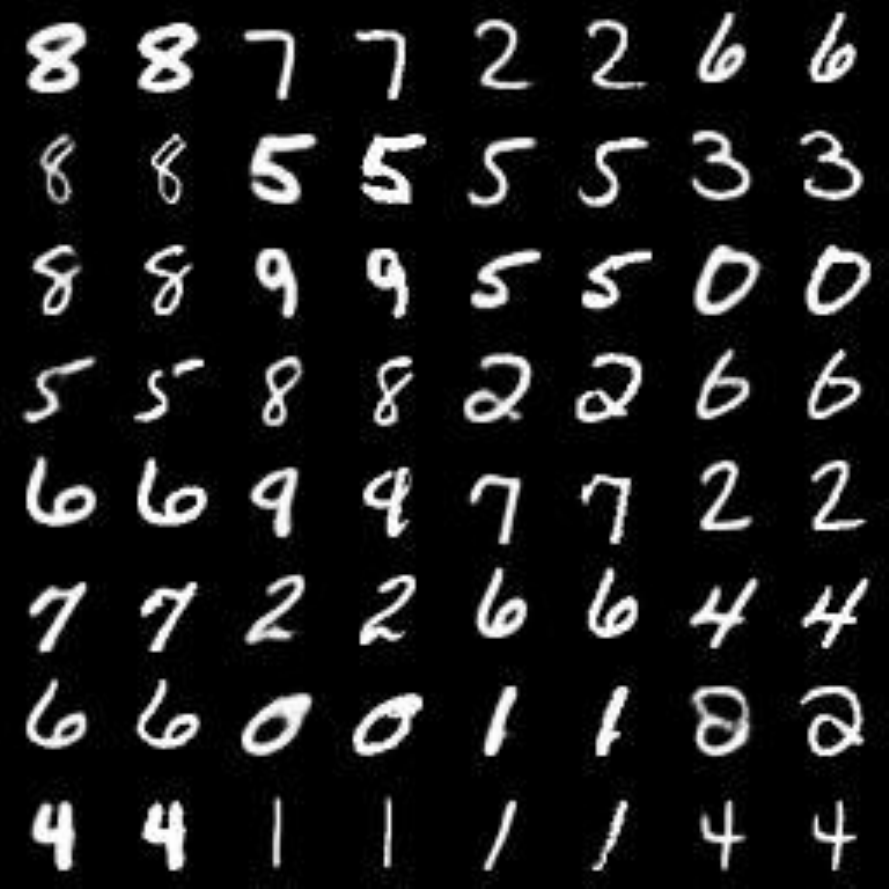}
    }
    \subfigure[CELEBA reconstruction.]{
    \includegraphics[width=0.31\linewidth]{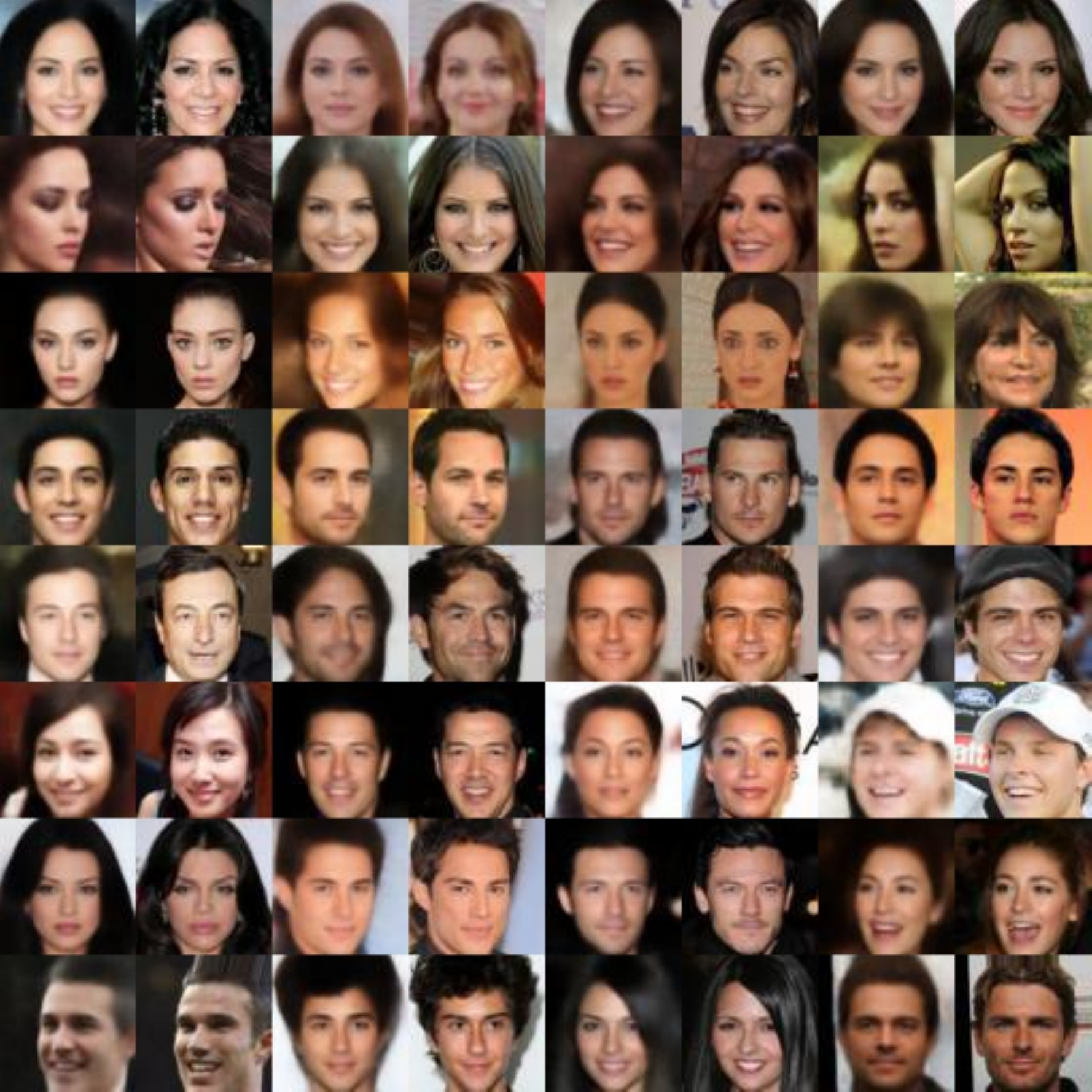}
    }
    \subfigure[CIFAR10 reconstruction.]{
    \includegraphics[width=0.31\linewidth]{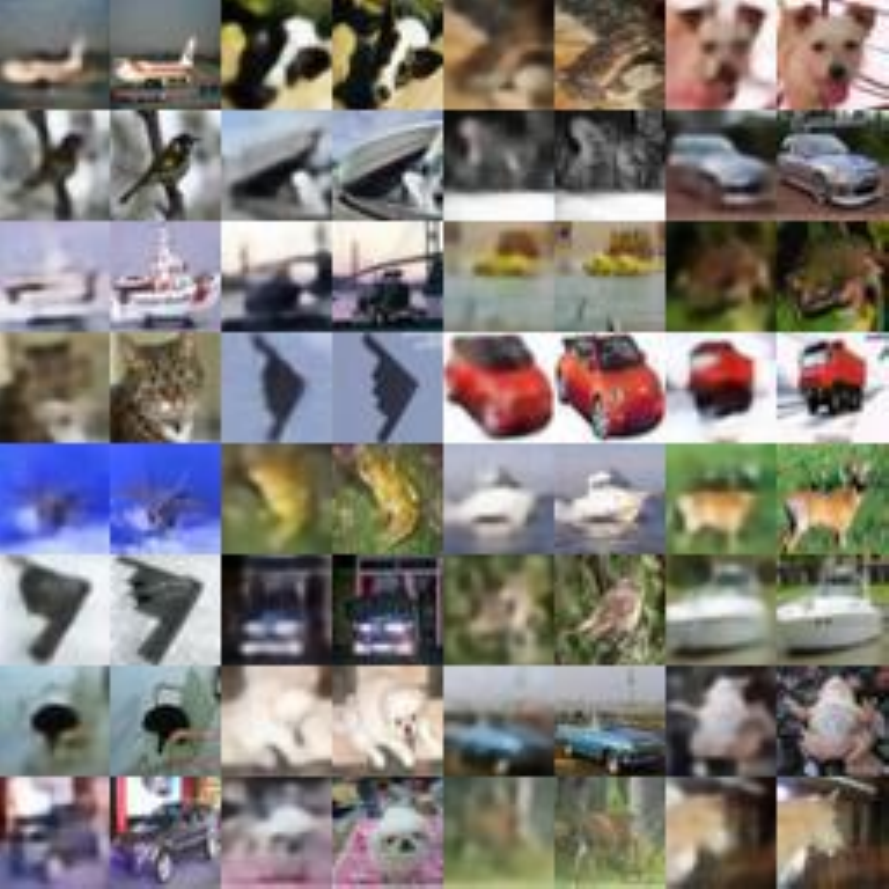}
    }
    \subfigure[MNIST sample.]{
    \includegraphics[width=0.31\linewidth]{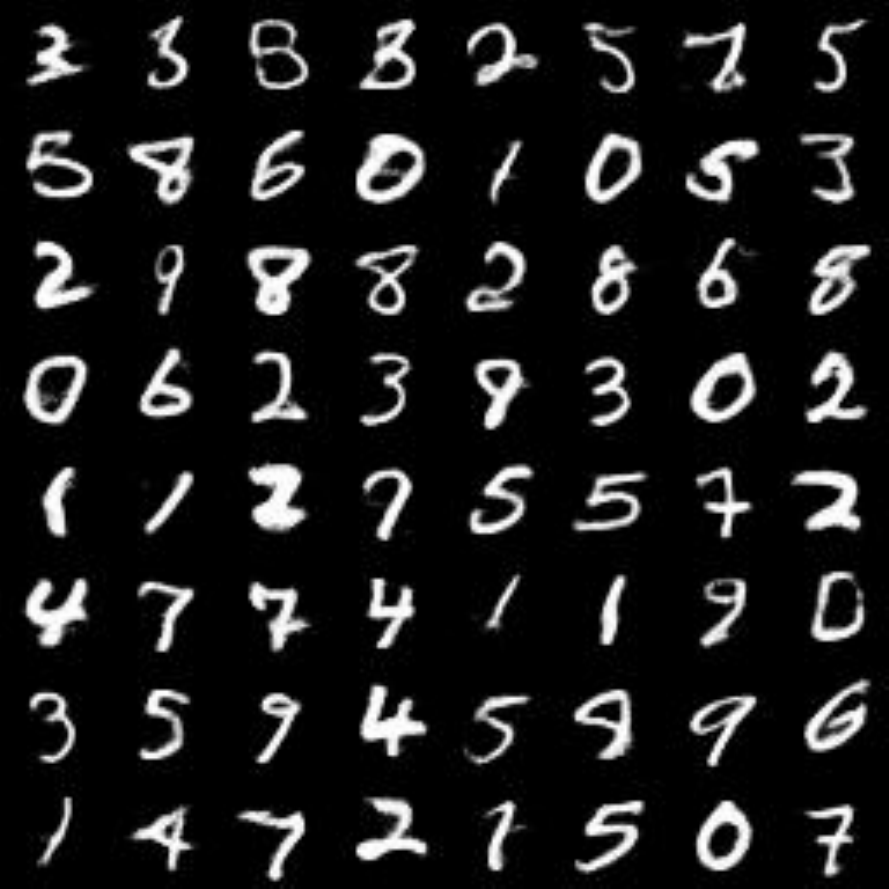}
    }
    \subfigure[CELEBA sample.]{
    \includegraphics[width=0.31\linewidth]{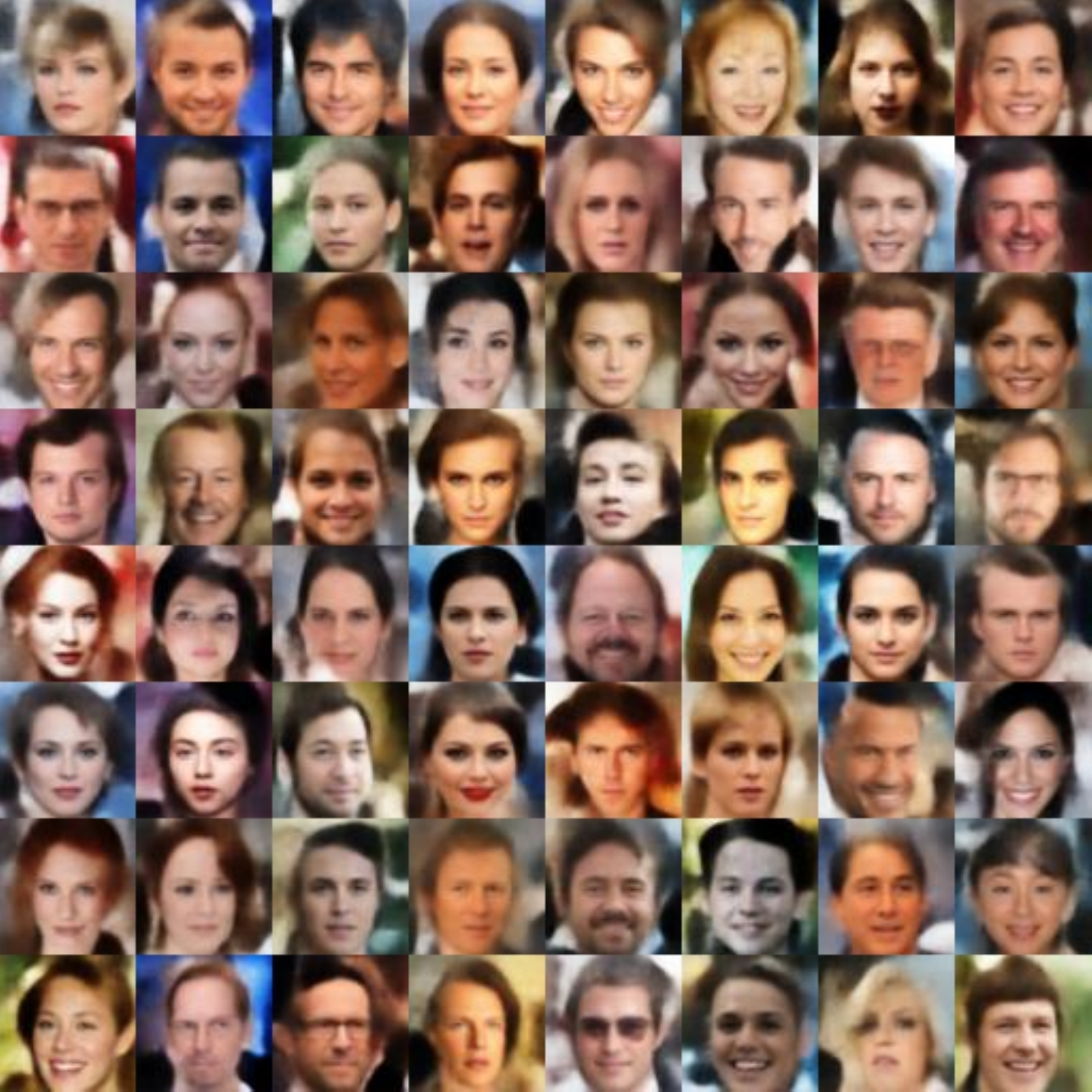}
    }
    \subfigure[CIFAR10 sample.]{
    \includegraphics[width=0.31\linewidth]{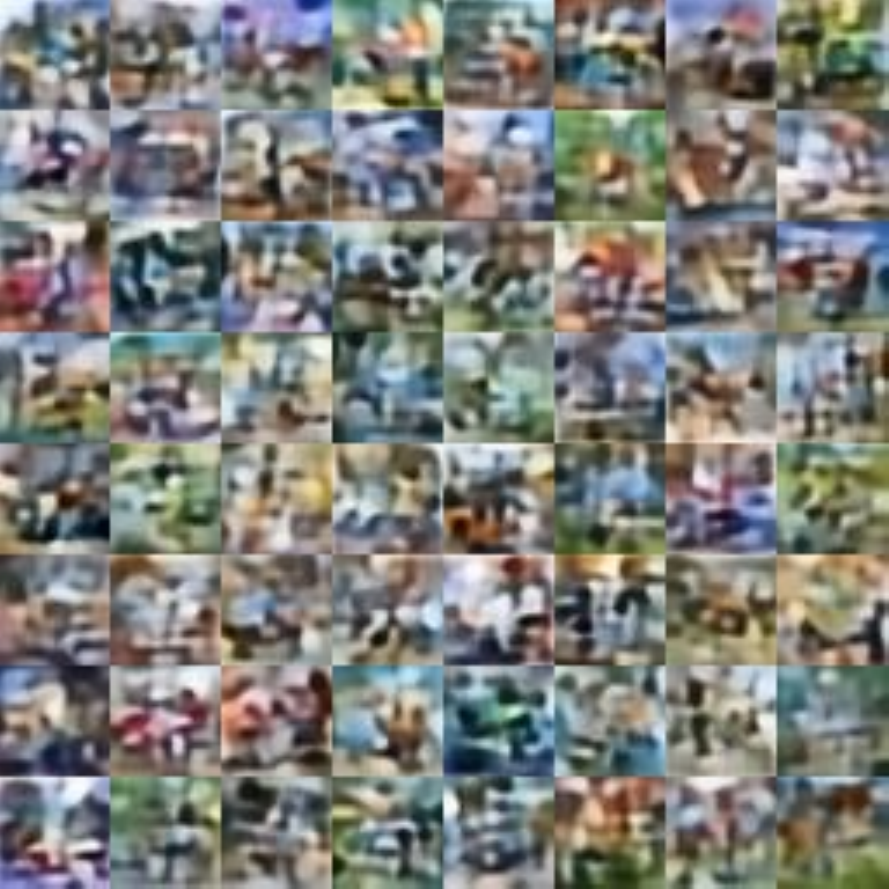}
    }
    \subfigure[MNIST interp.]{
    \includegraphics[width=0.31\linewidth]{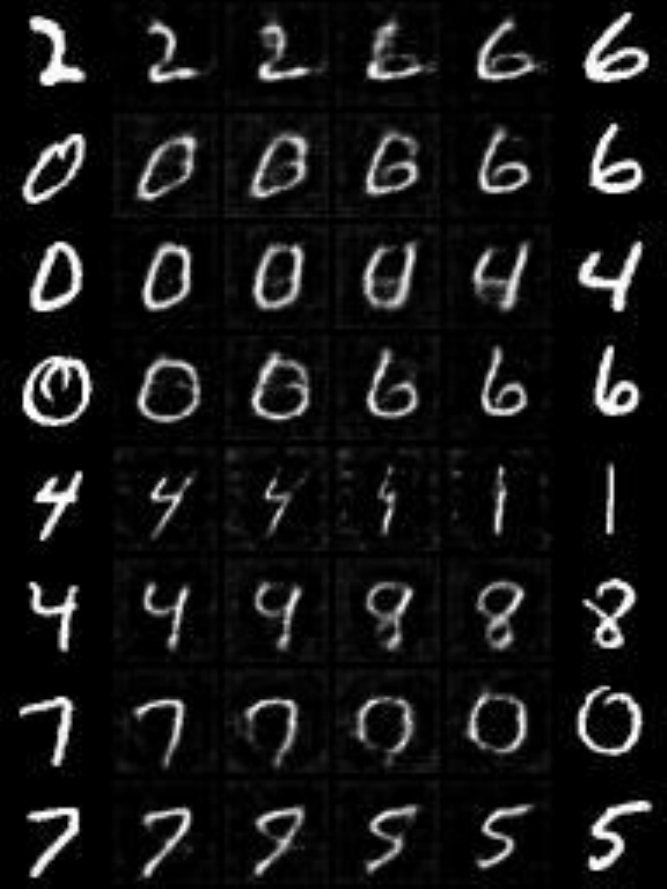}
    }
    \subfigure[CELEBA interp.]{
    \includegraphics[width=0.31\linewidth]{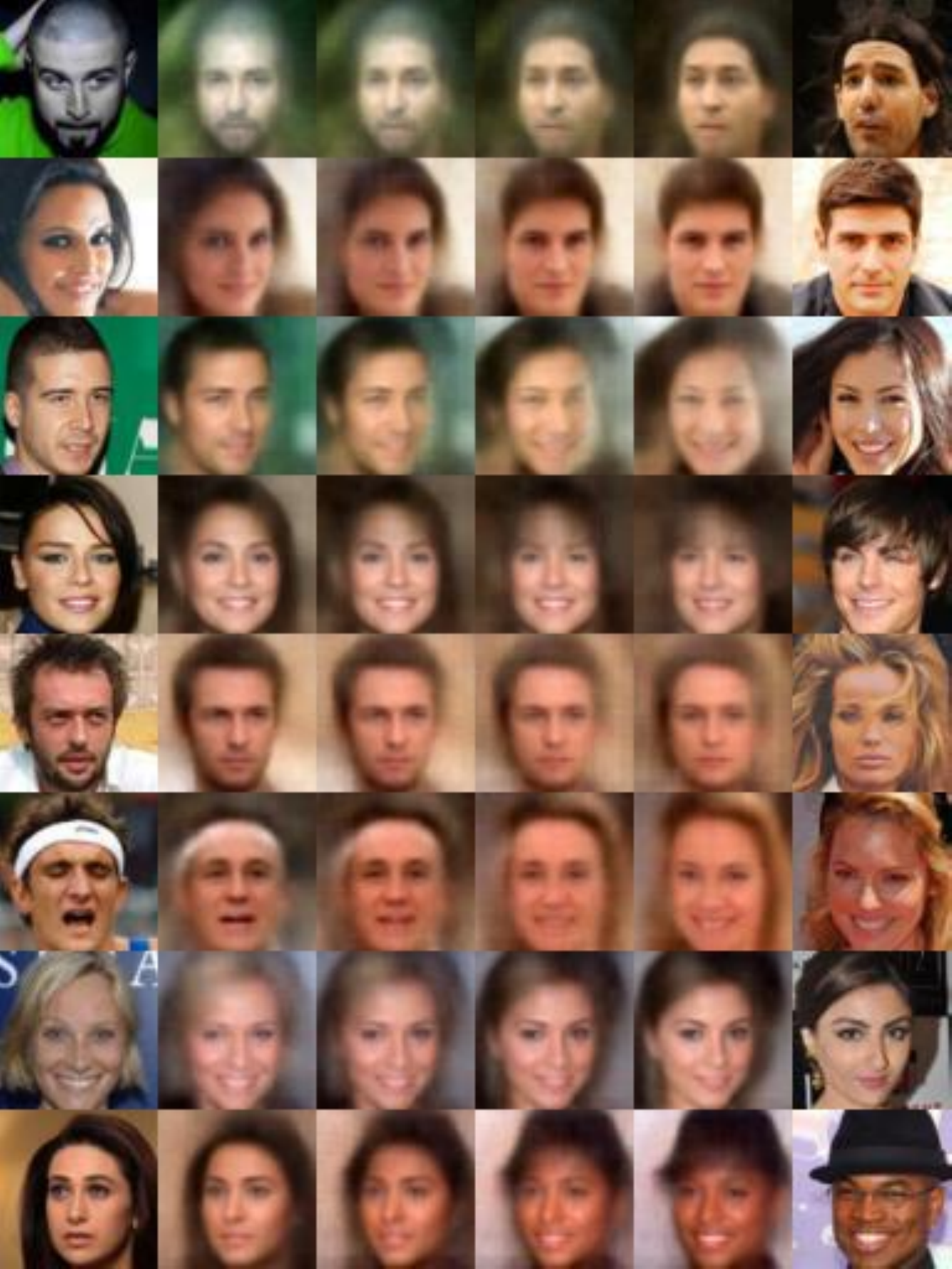}
    }
    \subfigure[CIFAR10 interp.]{
    \includegraphics[width=0.31\linewidth]{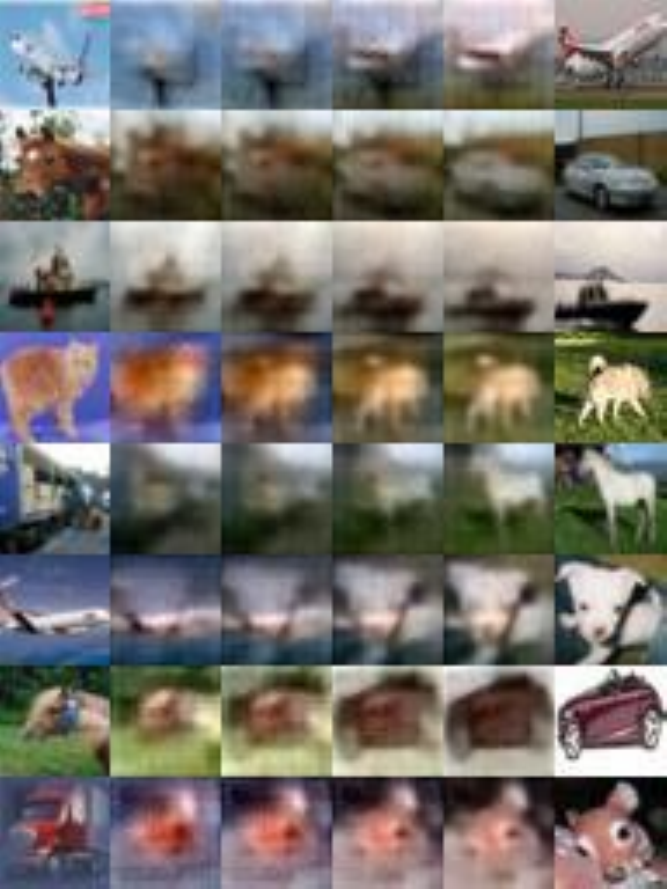}
    }
    \caption{The qualitative results for image reconstruction, generation and interpolation on MNIST , CELEBA and CIFAR10. We set the odd columns for reconstruction images and even columns for original images in all the reconstruction results. For the sample results, we use the sampler net to sample from the Gibbs posterior to generate the new images. And for the interpolation results, we use the random pairs image from the test set and compute their latent variable through the encoder. Every row in the interpolation results are the weighted mean of a pair of latent variable. The first and last columns in the interpolation results are set to the original test set image, with their interpolation results in the middle.}
    \label{rec_sample}
\end{figure*}

For quantitive evaluation, we record the FID score \cite{heusel2017gans} on MNIST \cite{lecun1998gradient}, CELEBA \cite{liu2015faceattributes} and CIFAR10 \cite{krizhevsky2009learning} of our model. Researchers commonly adopt these dataset in academia and industry to evaluate VAE. To make a fair comparison and prove the effectiveness of our method, we choose vanilla VAE and several improved versions which have been reported in the same network architecture. 
The baseline models include various VAE, namely constant-variance VAE (CV-VAE) \cite{balle2016end, ghosh2019resisting}, Wasserstein VAE (WAE) \cite{tolstikhin2017wasserstein} , 2-stage VAE (2s-VAE) \cite{dai2019diagnosing} and Regularized AutoEncoders (RAE) \cite{ghosh2019variational}. 

Specifically, The CV-VAE improved the vanilla VAE by fix the variance of $q_{\theta}(z|x)$, these leads to a stochasticity reduction for the encoder distribution. The WAE however, optimizing the VAE using two different regularizer, adversarial loss (WAE-GAN) or maximum mean discrepancy (WAE-MMD). We didn't choose the WAE-GAN for comparison, since it uses the adversarial loss which leads to an unstable training. The 2-stage VAE uses a two-stage remedy to learn a relatively precise latent space. And for RAE, they propose a deterministic training strategy which can bypass the optimization of ELBO. We compare with it since RAE can complete the same purpose as LDC-VAE do.

As can be seen in Table \ref{FID score}, attribute to the latent space consistency, our LDC-VAE outperform all the comparison models on MNIST, reducing the FID score form 10.03 to 1.02 and 8.69 to 4.73 for reconstruction and sample respectively. Also for the interpolation, our FID is much more lower than the secondary (8.09 for our model and 14.34 for WAE-MMD). On CELEBA, LDC-VAE achieves superior sample performance, i.e., the FID score is improved form 40.95 to 36.84. Besides, for reconstruction, LDC-VAE performs secondary performance that almost equals to the best one (i.e., WAE-MMD). On CIFAR10, our model obtain a similar level of performance as the current advanced VAE models.

For qualitative result, we show the reconstruction, sample and interpolation results on all the three datasets in Figure \ref{rec_sample}. As is shown in the result, our model have good reconstruction performance on all the three datasets, and can generate clear image on MNIST and CELEBA. The interpolation result shows that our method makes a reasonable model in the latent space.

\section{Conclusion and Future Works}
We propose the latent distribution consistent VAE (LDC-VAE), which serves as an alternative way to bypass the problems in ELBO. Firstly, we discuss inconsistency in the latent space of VAEs that are related to the ELBO, and introduce our LDC-VAE to avoid them. According to our derivation, the VAEs can be trained in a ``posterior to posterior" strategy and lead to the consistency in latent space theoretically, which are supported by our latent space visualizing experiment. Also, by applying our method, our LDC-VAE can achieve best performance in MNIST than several advanced VAEs. And in CELEBA and CIFAR10, our model is also comparable to several advanced VAEs. The future works will include the exploration of variant sampler net architecture which can obtain better sampling  performance especially in high dimensional.

\bibliography{IEEEabrv,aaai22.bib}

\end{document}